\documentclass[letterpaper]{article} 
\usepackage{aaai2026}  
\usepackage{times}  
\usepackage{helvet}  
\usepackage{courier}  
\usepackage[hyphens]{url}  
\usepackage{graphicx} 
\usepackage{natbib}  
\usepackage{caption} 
\usepackage{algorithm}
\usepackage{algorithmic}

\usepackage{newfloat}
\usepackage{listings}
\DeclareCaptionStyle{ruled}{labelfont=normalfont,labelsep=colon,strut=off} 
\floatstyle{ruled}
\newfloat{listing}{tb}{lst}{}
\floatname{listing}{Listing}
\title{Agentifying Agentic AI\footnote{In: WMAC 2026 - AAAI 2026 Bridge Program on Advancing LLM-Based Multi-Agent Collaboration}}
\author{
    Virginia Dignum, Frank Dignum
}
\affiliations{
         Umeå University\\
    virginia@cs.umu.se, dignum@cs.umu.se

}

\usepackage{bibentry}

\begin{document}

\maketitle

\begin{abstract}
Agentic AI seeks to endow systems with sustained autonomy, reasoning, and interaction capabilities. To realize this vision, its assumptions about agency must be complemented by explicit models of cognition, cooperation, and governance. This paper argues that the conceptual tools developed within the Autonomous Agents and Multi-Agent Systems (AAMAS) community, such as BDI architectures, communication protocols, mechanism design, and institutional modelling, provide precisely such a foundation. By aligning adaptive, data-driven approaches with structured models of reasoning and coordination, we outline a path toward agentic systems that are not only capable and flexible, but also transparent, cooperative, and accountable. The result is a perspective on agency that bridges formal theory and practical autonomy.
\end{abstract}


\section{Introduction}
The growing enthusiasm for Agentic AI reflects a long-standing ambition in Artificial Intelligence to build systems that are able to act autonomously, reason purposefully, and interact meaningfully with others (agents or persons). Yet, despite the novelty of current approaches, mostly based on large language models (LLMs), the challenges Agentic AI faces are not new. Questions of autonomy, coordination and accountability have been at the heart of Autonomous Agents and Multiagent Systems (AAMAS) research for decades now. What has changed is the context in which Agentic systems are developed: large-scale data-driven systems aiming to act in open environments. However, the same socio-technical challenges of autonomy, interaction, and adaptability remain. 

Current agentic systems promise flexible reasoning and continuous action, but often lack the explicit architectures, communication and reasoning semantics, and normative grounding that earlier agent models provided. Conversely, AAMAS research developed rich, explicit models of cognition, communication, and social structure, but often lacked the flexibility and scalability that characterize current data-driven approaches.

This paper argues that in order to make agentic AI systems truly agents, learning-based mechanisms must be complemented by structured reasoning and coordination models. Concepts such as reasoning architectures, formal interaction protocols, norms, and institutional governance provide the conceptual tools needed to turn behavioural autonomy into \textit{responsible} agency. By aligning and complementing the adaptive power of foundation models with the explicit structure of agent-based reasoning, we outline a path toward systems that are not only capable and flexible, but also coherent, cooperative, and accountable.\\

When people discuss the combination of LLMs and agents they usually start from the premiss that LLMs are tools that bring a lot of "intelligence" that should be harnessed by the agent. A (simplified) comparison between traditional agents and multi-agent systems and Agentic AI would say that traditional agents are controlable but limited. Due to the fact that all intelligence needs to be added manually, they might miss the knowledge to act in certain situations. Agentic AI is based on the power of LLMs that is known to be able to generate decisions in all situations, but might sometimes do very unexpected and harmful things.\\
From the view point of computer science, the latter is very undesirable. It prevents the agents to be verifiable and validated. That is, the function of the module and its   inputs and outputs cannot be \emph{guaranteed}. In a proper design of a system these guarantees are needed to build a complete system of which one can later on show that it lives up to expectations and that can be verified in some way.\\
Our argument is that LLMs are not easily seen as such predictable and specifiable components. There are two aspects of LLMs that make them inherently more difficult to handle.\\ First LLMs imitate human intelligence in an input-output way. I.e. they can mimic people in many cases by producing output (text) that people would produce given certain input. That does not mean that the LLM is doing the same or that you can expect it to do something in the future based on your experienced interactions with it in the past. In other words the results from the past are no guarantee for results in the future. E.g. a question to book flights to go to a conference might result in a booking of a morning flight to go there and arrive just in time for the first session and leave just after the last session finished. However, the next time it might book a return flight in the morning of the last day of the conference because it happens to be on sale. Of course this could be remedied by careful prompting. But just like with traditional agents, if the user forgets something you will get bad results that are not foreseen. For some types of systems this might not be a big problem, but for systems that are designed to achieve certain predefined goals or targets that is pretty bad. In some sense producing bad results or performing unwanted actions is worse than not performing any action at all. If an agent is part of a multi-agent system other agents might start acting autonomously based on this unwanted result and the user might only notice many steps later that something is wrong. This means that potentially a lot of actions have to be undone and one needs to sort out which those were and what went wrong in the first place. If an agent cannot take an appropriate action the user will be prompted and can solve the issue right away without elaborate roll backs being necessary.\\
The second problem of using LLMs is that they "live" in a world of data. If we would feed an LLM a billion documents stating that 2+2=5 it will start to replicate that. The fact that LLMs do make sense most of the time is purely based on the fact that many data is produced by sensible people and that creates an average of sensible outputs. However, this is less the case in contested areas like climate change or the utility of vaccinations. Although scientists might agree upon the facts, there are many people for whom the truth is not convenient and if they produce enough data the LLMs might start following the majority of the data produced instead of the most sensible or scientific data. I.e. the truth belongs to those who scream loudest.\\
The second problem is especially problematic if the LLM is used in an agent that tries to achieve something in the real world. Because the data it uses to base its actions on are not always properly linked to the real world and the actions it tries to perform might be possible on the data, while not possible or very undesirable in the real world. A good example of this is when an Agentic AI needs to make an appointment with a person, the person is not available and therefore it makes an appointment with someone else who is available (instead of trying different times or days).\\
The above issues lead us to the conclusion that although Agentic AI is appealing from a point of being very robust, it can also be very unreliable. Therefore the way to combine agents and Agentic AI might be to start with traditional agents and then fill in specific parts by LLMs. E.g. letting the LLM take care of practicalities of executing a complex action. The agent can check that the right (expected) result is achieved while the LLM makes sure that there always is some result.

Although the above conclusion is already an interesting one to start with, it only considers the actions of a single agent. This follows most of the current discourse on agentic AI focuses which also focuses narrowly on single agents. A framing that seems to ignore the fact that any meaningful environment already contains other agents, human, artificial, or institutional, with their own objectives, values, and constraints, and therefore the main aspect of agents should be to navigate this multi-agent environment (even if simply by ignoring all other agents).\\
This single-agent focus leads to individualistic AI models: optimizing behavior in isolation rather than reasoning about interaction, negotiation, coordination, and shared norms.
Progress in ‘agentic AI’ should start by looking at what has been done in the AAMAS community 
for recent work and \cite{weiss2013multiagent} for an overview of earlier work) for several decades now. We fully accept and acknowledge that the AAMAS community has failed to communicate and leverage their own work but still there are many usable results to build upon. In the table in figure 1 below, we provide a short overview of issues that are of great use to advance ‘agentic AI’ in a multi agent setting.

We argue that true agency only emerges in relation to others. Therefore ignoring multi-agent dynamics leaves critical gaps in safety, accountability, and governance, since many risks stem not from an agent acting alone, but from agents acting together, or at cross-purposes, without mechanisms for communication, alignment, trust, and collective oversight.
To align with a more realistic and responsible conception of agency, the focus should shift from what an agent can do to how agents coexist, coordinate, and co-regulate.

Incorporating insights from multi-agent systems research, on roles, norms, organizational structures, and verified communication, would turn the landscape from a catalogue of architectures into a framework for designing agentic ecosystems. These are not just technical challenges but socio-technical ones, requiring governance, interaction protocols, and institutional embedding.


In our view, true benefit of ‘agentic AI’ starts when we see these not as defined by the autonomy of individual systems, but by their capacity to operate within and contribute to a shared social context. Without this relational dimension, ‘agentic’ remains a misnomer, it refers to ‘autonomous AI’ without capacity to address agency.\\
In the rest of this paper we will discuss the multi agent aspects mentioned in figure 1 and show how they could be used for Agentic AI and more importantly why some of these aspects are difficult to implement in Agentic AI as long as these are black box systems that might show realistic behavior but cannot be reasoned about.



\section*{Agentic AI Systems}
Agentic AI denotes systems that couple large-scale foundation models with capabilities to \emph{reason}, \emph{act} (e.g., via tools or environments), and \emph{interact} with users and other systems in a sustained, goal-directed manner.
While this shift extends AI from passive prediction to task-oriented autonomy, it also surfaces open problems, as indicated in recent analyses, such as the AAAI 2025 Presidential Panel on the Future of AI Research \cite{aaai2025panel}.  These challenges include:
\begin{enumerate}
  \item[(i)] \textbf{Reliability and grounding}~\cite{plaat2025agenticllm}. 
  Agentic systems often struggle to connect their language-based reasoning with the actual state of the world. They may generate plausible but false information, or take actions that do not reflect real conditions, particularly when using tools or interacting with digital environments. This lack of grounding can lead to cascading errors and undermines user trust.

  \item[(ii)] \textbf{Long-horizon agency}~\cite{zhang2024memorysurvey}. 
  Although these systems can plan or reflect in short sequences, they remain fragile when tasks require sustained reasoning, long-term goals, or adaptation over time. Their internal memory and self-monitoring are still rudimentary, which limits their capacity for persistence, learning from past outcomes, and maintaining coherent strategies.

  \item[(iii)] \textbf{Evaluation}~\cite{aaai2025panel,aisiEvalApproach2024,aisiFourth2024}. 
  There are no agreed methods or benchmarks to assess the performance, safety, or autonomy of agentic systems. Current evaluations are fragmented, often focusing on narrow tasks or short interactions. As a result, it is difficult to compare systems or determine whether an agent is genuinely capable, reliable, or aligned with intended goals.

  \item[(iv)] \textbf{Risk management and governance}~\cite{nistrmf100-1,nistgenai600-1}. 
  Once deployed, agentic systems can act continuously and independently, which raises unresolved questions about accountability, oversight, and liability. Few governance frameworks clearly specify who is responsible for an agent’s actions, how its decisions can be audited, or when human intervention should occur. Transparency and safety assurance remain limited.

  \item[(v)] \textbf{Security and privacy}~\cite{aisiResearchHub2024,whittaker2025privacy}. 
  The ability of agentic systems to access tools, data, and online services introduces new vulnerabilities. Malicious inputs, excessive permissions, or unintended data sharing can expose sensitive information or allow agents to be manipulated. Ensuring security and privacy therefore requires strict controls on access, monitoring, and data handling.

  \item[(vi)] \textbf{Value and maturity}~\cite{zohaib2025agenticmaturity,gartner2025reuters}. 
  Despite significant enthusiasm, the practical benefits of agentic systems are still uncertain. Many projects remain experimental, face high maintenance costs, or fail to deliver stable value once deployed. Expectations often exceed current technical maturity, which risks both wasted resources and loss of confidence if results do not meet claims.

  \item[(vii)] \textbf{Costs vs. benefits}
  If the Agentic AI makes use of a standard foundation system, every time it accesses that system it will incur a cost. It does not seem a lot, but can quickly amount to a few hundred dollars per month! A traditional agent system has neglectable running costs compared to this. Moreover, being tied to a an API of a big tech company is a cost in itself as well. Creating your own foundation model is an alternative, but that might incur large development costs.
\end{enumerate}

Taken together, these challenges reveal that current agentic systems, while powerful, lack conceptual and structural coherence.  
Their autonomy is largely behavioral rather than reasoned, and their capacity for coordination, explanation, and accountability remains limited.  
The following section examines how established concepts from the AAMAS tradition, such as explicit reasoning architectures, communication protocols, and social structures, offer perspectives and methods that can help address these open issues.

\section*{Bridging AAMAS and Agentic AI: Conceptual Comparison and Research Foundations}

The field of \emph{Autonomous Agents and Multi-Agent Systems (AAMAS)} has, for more than three decades, provided the conceptual and formal underpinnings of what it means for artificial entities to act autonomously, reason about their environment, and interact with others.  
AAMAS research explores how individual and collective intelligence can emerge from explicit representations of beliefs, goals, commitments, and social relations, supported by logic-based reasoning, coordination protocols, and organizational structures.  
Its focus extends beyond technical autonomy to include explainability, accountability, and cooperation—features that define agency not merely as the capacity to act, but as the capacity to act \emph{appropriately} within a social and institutional context.

\begin{figure*}[!t]
\centering
\includegraphics[width=\textwidth]{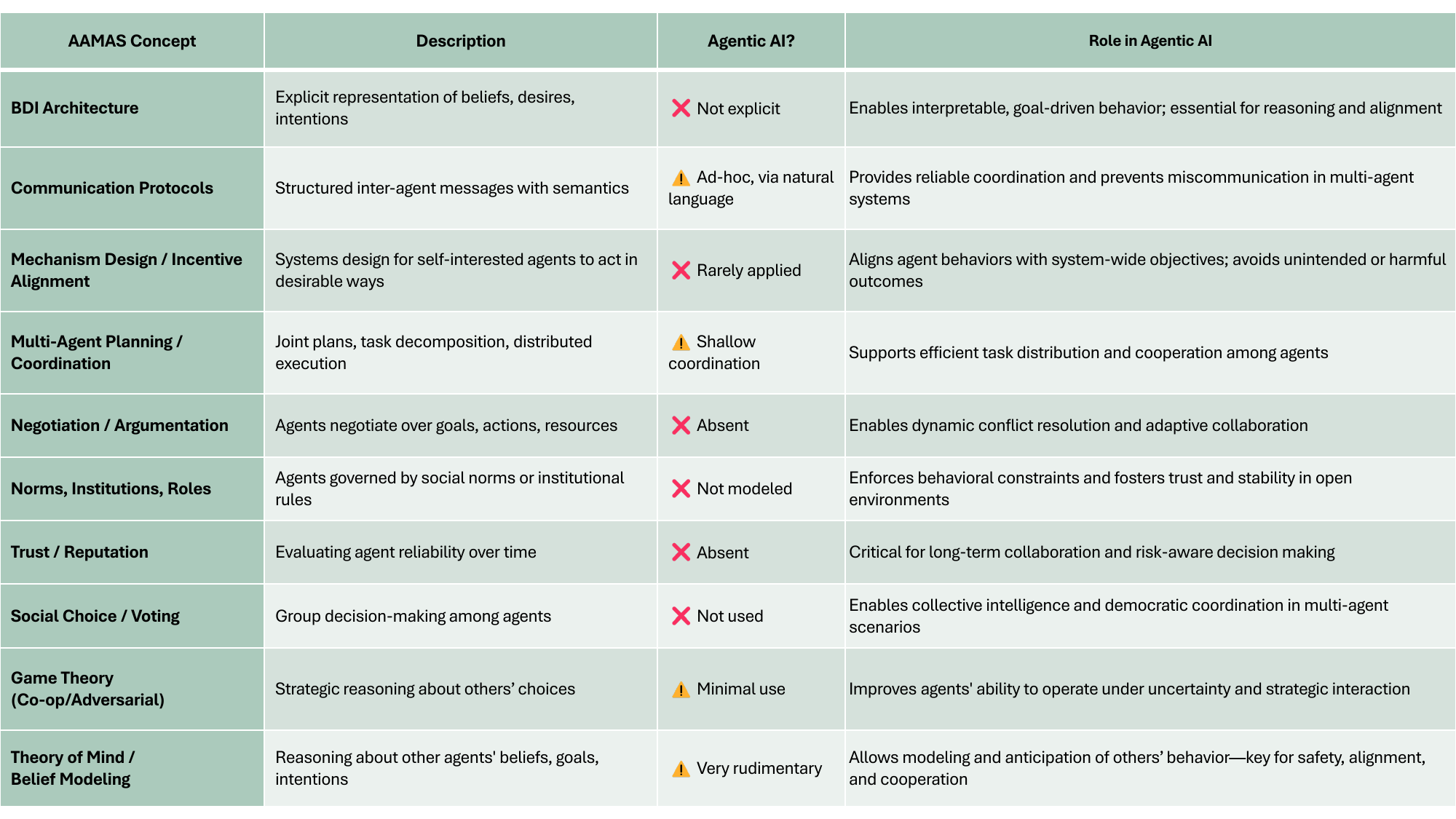} 
\caption{Conceptual comparison between classical AAMAS foundations and emerging Agentic AI paradigms. 
The figure summarizes how established research in Autonomous Agents and Multi-Agent Systems (AAMAS) provides structured and interpretable foundations for the design of responsible and cooperative ``agentic'' AI systems. Each row contrasts a well-understood AAMAS concept with its current realization, or absence, in contemporary Agentic AI approaches.}
\label{fig:aamas-agentic}
\end{figure*}

Figure~\ref{fig:aamas-agentic} highlights how the theoretical and methodological foundations developed within the AAMAS community remain profoundly relevant to current discussions on Agentic AI.  

While AAMAS research approached agency as a structured, explicit, and explainable process, grounded in formal models of reasoning, communication, and coordination, contemporary Agentic AI tends to emphasize large-scale learning and behavioral emergence without comparable levels of conceptual or formal clarity.   
This contrast is not merely historical or terminological: it reflects a shift from explicitly modelled cognitive and social processes to implicitly inferred ones, with significant implications for transparency, accountability, and control.
Rather than risking reinventing the wheel, Agentic AI could benefit from building on the AAMAS tradition of structure, justification, and verifiability to address its current challenges of coherence, coordination, and responsible behavior, which provides stronger foundations for agentic systems capable of reasoning, collaboration, and ethical accountability.

In the remainder of this section, we outline the main concepts and results developed within the AAMAS tradition and discuss how they provide theoretical and methodological tools to understand, and potentially overcome, the current limitations and challenges faced by emerging Agentic AI systems.

\paragraph{Explicit Architecture. }
Classical agent theories such as the \emph{Belief--Desire--Intention (BDI)} model~\cite{rao1995bdi, wooldridge2009introduction} exemplify the structured and explicitly formal architectures that characterize AAMAS research.  
The BDI paradigm models agency through distinct mental components: \emph{beliefs} represent the information an agent has about the world; \emph{desires} capture its motivational or goal-oriented states; and \emph{intentions} encode the subset of desires that the agent has committed to pursue.  
This tripartite structure provides a transparent and verifiable mapping between perception, deliberation, and action, supporting predictable, explainable, and goal-consistent behaviour. Most important is that the autonomy that an agent has to perform actions and plan and re-plan are limited by the explicit goal that the plans and actions are trying to achieve. Thus it is possible to reason about when a goal should be given up as it is deemed to be unachievable (possibly due to a changing environment). E.g. when an agent tries to book a flight for a person to a certain destination and date(s), it can stop trying after it looked for all possibilities of direct and indirect flights going in a certain spatial corridor to the destination. Thus it is prevented from trying to book a flight from Amsterdam to Paris that goes via Rome. Although Agentic AI systems can be given some restrictions in their prompt it is well known that they often do not cope very well with combinations of numerical constraints.  

Within AAMAS tradition, such architectures are not merely conceptual but serve as a foundation for reasoning about autonomy, cooperation, and accountability across distributed agents. E.g. in applications where agents have to cooperate to achieve a goal they might want to reason about what information another agent has available in order to communicate additional information that is needed to achieve their goal. E.g. if two agents try to book flights for colleagues to the same conference they might want to communicate the flight number and seat number such that the other agent can try to book the same flight and a seat that is adjacent.

Thus having explicit architectures with mental attitudes enable explicit introspection, communication of mental attitudes, and logical verification of consistency between goals and actions, features that make systems both interpretable and governable.  
By contrast, contemporary Agentic AI systems, particularly those built upon large language models, typically operate without explicit internal state representations or commitment mechanisms.  
Their ``intentionality'' is implicit, inferred from surface-level behavior rather than formally specified reasoning.  
As a result, such systems may appear adaptive but lack the structural guarantees that link motivations to actions, limiting their alignment, predictability, and verifiability.

\paragraph{Communication Protocols.}
Communication has always been central to multi-agent systems.  
Classical AAMAS research developed formal languages such as KQML and FIPA-ACL~\cite{finin1994kqml}, which define not only message syntax but also the semantics of communicative acts like requests, commitments, or promises.  
These frameworks ensure that agents can exchange information, negotiate, and coordinate actions with a shared understanding of intent and context.  
In contrast, most Agentic AI systems rely on unstructured natural language dialogue, mediated by large language models.  
While flexible and expressive, such exchanges lack guaranteed semantics: meaning is inferred statistically rather than interpreted formally. E.g. suppose I know that someone is arriving tomorrow at university and I have reasons to try and avoid them. I do not know when the person arrives exactly which makes it difficult to avoid them. If my personal agent asks the personal agent of the host whether it knows the arrival time of the guest, I just want that agent to state the arrival time if it can look it up in its own information. I would not like it to start asking the guest for the arrival time on my behalf (giving my request as the reason for the question). In traditional agent systems one can distinguish these two situations by sending different types of requests for information. Thus avoiding potential awkward situations. 
The natural language communication of agentic AI does not use this strict semantic approach and can thus lead to misunderstanding, inconsistent commitments, or spurious alignment between interacting agents.  
Reintroducing formal semantics or constrained communication layers could therefore improve reliability, traceability, and accountability in multi-agent interactions.

\paragraph{Mechanism Design and Incentive Alignment.}
A central insight of the AAMAS community has been that effective collective behavior requires not only intelligent agents but also mechanisms that align individual incentives with group objectives.  
Building on game theory and social choice, mechanism design provides mathematical tools for structuring rewards, penalties, and information flows so that rational agents act in ways that support system-level goals, even when they pursue their own objectives. AAMAS research has developed a rich body of work on distributed mechanism design, coordination protocols, and incentive engineering for autonomous agents~\cite{roughgarden2010algorithmic, sandholm2003automated, wellman2006decentralized, zlotkin1996cooperation}.  
These approaches specify how agents with potentially conflicting preferences can interact strategically while ensuring outcomes that are efficient, fair, or socially desirable. They capture the essence of multi-agent rationality: designing the environment, not the agent, to guarantee cooperative equilibria and stable outcomes. A good example where this mechanism design works very well is in logistics. In an environment where agents represent trucks with different capacities and characteristics and packages are offered dynamically to the system at different places and with different destinations, and auction mechanism where agents bid on packages is a very robust and efficient way to get all packages delivered. Each agent can calculate how much it would cost them to deliver the package (depending on their current position and already planned route). The agent with the lowest cost wins the package.\\
Essential is that the intelligence of finding solutions is divided between the mechanism and the agents. One can easily add more agents to the system or take them away. Agents also can have there own objectives and intelligence. E.g. they can learn from past experience that new packages are offered regularly at some places and anticipate on that.\\
In the other hand different overall system objectives can be handled easily. E.g. one might want to optimize the delivery time of packages by either optimizing the average delivery time. Or minimize the slowest delivery time of packages. Or one could minimize the fuel consumption of the whole fleet. 

By contrast, most contemporary Agentic AI systems treat agency as an essentially \emph{individual} property~\cite{gao2025singleagentmultiagent,sapkota2025aiagentsagentic}.  
Their design and evaluation typically concern how a single agent plans or acts autonomously, rather than how multiple autonomous entities coordinate, negotiate, or balance incentives in shared environments.  
This single-agent orientation explains why interaction mechanisms, incentive structures, and multi-agent dependencies are often overlooked.  
As a result, agentic systems may exhibit emergent competition, goal conflict, or inefficient collaboration once deployed in social or multi-agent settings.  
Reintroducing mechanism design principles, i.e. the explicit modeling of preferences, incentives, and interaction rules, could help ensure that autonomous agents pursue goals that remain coherent, mutually compatible, and aligned with collective human and institutional objectives.

\paragraph{Multi-Agent Planning and Coordination.}
AAMAS research has long investigated how multiple agents can plan and act together through frameworks such as joint intention theory, distributed planning, and coalition formation~\cite{grosz1996collaborative, durfee2001distributed,lesser2002evolution}.  
These approaches define how agents share goals, decompose complex tasks, and maintain mutual awareness during execution. For a large part these frameworks make it possible to balance efficient central control and robust local adaptations (see also \cite{agentbasedcrisismanagement}). The difficulty in multi agent planning is that the agents have to work together to achieve a common goal, while the agents also should be autonomous enough to solve local problems. In a multi agent plan an agent should be aware to some extent about the overall plan such that it knows which agents should be providing the input for its own task and which agents are dependent on its tasks. E.g. if there are two agents together booking a trip, where one agent books a flight and the other the hotel, they need to synchronize on the dates of the flight and the hotel booking. If the flight agent needs to extend the stay with one day because there are no flights available on the end day. Then the hotel agent should extend the stay in the hotel with a day. But this might not be possible in the favourite hotel and the hotel has to be changed. All of this should also be explained to the user and/or permission asked to make the changes. It shows that quite some monitoring and communication is needed in order to execute a plan properly. Especially when changes have to be made to the original plan.\\
In top of this one usually has to choose what type of agents to employ. Should all agents be as general purpose as possible or should they be specialized in specific tasks? Specialized agents can work far more efficient, but they need more coordination efforts with other specialized agents. Depending on the domain and how dynamic it is one needs more or less intelligence in the agents. Thus one adapts the intelligence and thus the complexity of the system to the domain. Do not make things more complex than needed. But complex enough to render the system robust and flexible enough to cope with changing contexts.

In contrast, current Agentic AI systems often exhibit shallow or emergent coordination: multiple agents may interact through language but lack shared models of goals, resources, or dependencies.  
As a result, their cooperation is opportunistic rather than structured, and their collective performance is difficult to predict or verify.  
Reintroducing formal models of joint planning could enable more coherent, robust collaboration among both artificial and human agents.

\paragraph{Negotiation and Argumentation.}
Negotiation and argumentation have been central to the study of decision-making in open multi-agent systems~\cite{rahwan2009argumentation, jennings2001automated}.  
These frameworks allow agents to reason about conflicting goals, justify their positions, and dynamically revise commitments through dialogue.  
They provide structured means of resolving disagreements, balancing interests, and reaching shared decisions.\\
Again the solutions are divided between a fixed mechanism that defines the rules for the negotiation or the argumentation and the intelligence of the agents that determines how they act and react within the context of the negotiation or argumentation framework. Negotiation is useful in contexts where the agents are competing and serving the interests of different parties. E.g. in the logistic domain one can negotiate about the price of transporting the furniture of a household. The furniture can be picked up right away but then the costs will be higher as the transporter has to store it a few days. Or it can be lower if the furniture can be picked up when transport to the destination is available. But in that case the owner has to keep the furniture somewhere (possibly having to keep his current house for another week or month).\\
Argumentation is important in cooperative settings where agents with different expertise try to find an optimal solution for a problem based on known facts. E.g. should an elderly patient have a treatment or surgery or not. What are the risks, what is the expected quality of life with and without the treatment, etc. Using arguments often brings to light new facts that were not known by at least some of the agents. E.g. there is a new surgical technique that is less evasive and reduces the risk of complication in surgery. This technique of argumentation only works if the agents can produce the causal links between the arguments and can reason about the support or refutation of arguments. This is something that Agentic AI is notoriously weak in doing.

Most current Agentic AI models lack explicit mechanisms for conflict resolution or deliberative exchange.  
Their coordination is largely cooperative by design, with limited capacity for structured debate, persuasion, or consensus building.  
Integrating argumentation-based reasoning could enhance the adaptability and accountability of multi-agent interactions, especially in socio-technical or policy-relevant domains.

\paragraph{Norms and Institutions.}
Social organization has long been a defining concern in AAMAS, leading to extensive research on how autonomy and coordination can coexist under shared rules and expectations.  
Norms, institutions, and organizational reasoning have long been central to AAMAS research on how autonomous agents can coordinate and remain accountable within collective contexts.  
Early work on normative multi-agent systems formalized how obligations, permissions, and prohibitions regulate autonomous behavior~\cite{boella2007norm}.  
Complementary research on \emph{Electronic Institutions}~\cite{esteva2001electronic, sierra2004eInstitutions} developed computational environments where interaction is structured by roles, protocols, and institutional rules.  
Organizational models such as \emph{OperA} and \emph{JaCaMo} provide frameworks for specifying how agents, roles, and organizational structures interact to achieve shared objectives~\cite{dignum2004model, boissier2013jacamo}.  
In parallel, formal approaches such as \emph{LAO} focus on the logical foundations of organizational reasoning, defining the semantics of roles, norms, and (re)organization processes~\cite{dignum2014formal}.  
These complementary perspectives, normative, institutional, organizational, and logical, demonstrate how autonomy can be embedded within explicit social and institutional contexts.  The comparative analyses presented in~\cite{alderweeld2016socialcoordination} review a range of frameworks that formalize social coordination through explicit roles, norms, and institutional rules.  
Together, these approaches model how agents’ autonomy is constrained and guided by collective expectations, linking individual reasoning to organizational and societal objectives.  
They make it possible to specify not only what agents \emph{can} do, but what they \emph{ought} to do within coherent systems of expectations, thereby enabling accountability and predictability in open multi-agent environments.\\
An easy implementation of all these social structures would be a set of constraints that drive and restrict the agents. However, this is fundamentally not how they function in human society and misses the crucial power of these structures. The importance of a norm becomes apparent not when all people follow it, but when people decide to violate it. Norms will be followed in normal circumstances. In these circumstances the norms provide expectations that simplify a complex context and interactions. But if a woman is about to give birth everyone accepts that her partner will speed to get her to hospital in time.\\
In the same way a boss can order an employee to perform some task. Normally the employee will obey, but only in the context of work related tasks that are within the role of the employee. E.g. you can ask a secretary to bring coffee to a meeting as they facilitate the meeting and this falls within that perimeter. However, you cannot ask them to get you coffee and bring it home to you.\\
These types of behaviours have been studied for many years in the agent community. The behaviours, and especially the violations, indicate an underlying shared value structure of a society that implicitly drives our behaviour in social contexts through the social structures. The very fact that violations do not occur often and have a wide variety means that they are very hard to detect by foundation models, even with large amounts of data. But more importantly, the current foundation models mainly use data from Western societies which do not adhere to the same value structures as other societies in the world.

Most current Agentic AI systems treat agents as self-contained entities operating without explicit normative or institutional constraints. The idea is that these social structures will be implicitly learned and expressed in the foundation model.  
However, the absence of explicit social structure makes it difficult to ensure consistent behavior, shared responsibility, or compliance with higher-level objectives.  
Embedding normative and organizational reasoning—capturing duties, expectations, and institutional context—would help bridge the gap between technical agency and societal governance, enabling agentic systems to operate as accountable members of human–AI collectives. E.g. agents that serve as personal assistants of humans should understand when to obey orders and when to question them. At this moment the guard rails (norms) are manually added and work only in a limited way. E.g. ask ChatGPT to tell a joke about fat people and it will deny doing this, giving a (standard) argument about why this is inappropriate. However, asking it to tell a joke about tall people it will give a (very bad) joke without problem. 

\paragraph{Trust and Reputation.}
In open and dynamic environments, agents must assess the reliability of others.  
AAMAS research developed computational models of trust and reputation to represent and update expectations about others’ behavior over time~\cite{sabater2005review}.  
These models enable selective cooperation, reputation-based incentives, and resilience to deception or error. E.g. if an agent is buying a plane ticket for its owner it should be aware about the reputation of the seller. Is the seller an intermediary that sells flights at suspiciously low prices then it might be that the seller is not planning to deliver after the ticket has been paid. Or it can be that there are many hidden costs that make the ticket more expensive than that of its competitors. Companies like Amazon that sell commodities are careful to keep a good reputation as that will draw customers. Therefore they make sure the ordered books are delivered in time and they will replace an order if it is damaged without problems. This kind of reputation is an important factor in choosing to deal with a company. Any agent representing a human customer should be aware of these aspects as well as it will influence the satisfaction of its customer.

However, most Agentic AI systems treat all interactions as stateless, lacking persistent memory of past performance or behavior. Thus they lack a mechanism to keep track of reputation and trust. Without mechanisms for trust accumulation and reputation management, agentic ecosystems remain vulnerable to exploitation, inconsistency, and loss of confidence.  
Incorporating explicit trust metrics could enhance stability, safety, and long-term cooperation among autonomous agents.

\paragraph{Collective and Strategic Decision-Making.}
A core strength of AAMAS research lies in its formal treatment of how multiple autonomous agents make decisions together, balancing individual goals with collective outcomes.  
Two complementary traditions underpin this work.  
First, \emph{social choice theory} provides formal methods for aggregating individual preferences, opinions, or utilities into collective decisions~\cite{conitzer2009computational, brandt2016handbook, endriss2014social}.  
Within AAMAS, these models have been extended to distributed contexts such as voting, resource allocation, coalition formation, and consensus building, ensuring that group decisions remain transparent, equitable, and strategically robust. The use of this social choice theory can already be illustrated with a simple calender agent that needs to coordinate days and times with other agents to determine the day and time for joint meetings. First one needs to determine the possible options by intersecting the available times given by all agents. An agent could game this part by giving only its preferred option, forcing the others to adopt that one if it overlaps with enough other options. This can be countered by asking for a minimum number of options. Then the voting for a possible option also has to be regulated. Can I gain insight by waiting very long before voting and seeing what the others voted already? Or should all votes be invisible until all have voted? All of these aspects have an influence in how efficient the calender system works for all users and part of it relies on the agents being aware of the mechanism, its user's preferences and ways to advance these preferences. 

\emph{Game theory} offers analytical foundations for reasoning about cooperation, competition, and strategic interaction among rational agents~\cite{osborne1994course}. Based on what are possible pay-offs of different outcomes for different players involved in an interaction one can make predictions about possible reactions to an action one takes. So, game theory makes use of a partial model of other agents (nl. their preference over outcomes of the interaction) to strategize on the best action.   
AAMAS research applies both cooperative and non-cooperative game theory to predict others’ actions, evaluate equilibria, and design mechanisms that align individual incentives with social welfare.  
Together, these approaches illustrate how autonomy and interdependence can coexist under principled models of collective rationality.  

Contemporary Agentic AI systems, however, rarely implement explicit aggregation or strategic reasoning mechanisms, even when deployed in multi-agent or human–AI environments.  
Their coordination tends to emerge implicitly from large-scale learning or heuristic adaptation, with limited grounding in preference aggregation, incentive alignment, or equilibrium analysis.  
Integrating social choice and game-theoretic reasoning could therefore strengthen the ability of agentic systems to deliberate, cooperate, and negotiate outcomes that are not only effective but also fair, stable, and aligned with collective human values.

\paragraph{Theory of Mind and Belief Modeling.}
Reasoning about other agents’ beliefs, goals, and intentions (often referred to as ``theory of mind'') has long been a cornerstone of cognitive and multi-agent reasoning~\cite{baker2017rational, deweerd2014agent}.  
Such modeling allows agents to anticipate others’ actions, cooperate under uncertainty, and interpret or justify observed behavior. It is used for strategic reasoning in game theory, negotiation and argumentation. It can very between using just some known preferences to being able to reproduce the reasoning of another agent.  
Within AAMAS, belief–desire–intention architectures and epistemic logic frameworks provide explicit formalisms for representing and reasoning about mental states of other agents. Given that you know the other agent uses a certain architecture with specific characteristics, you can make use of that knowledge to derive how an agent might react in a certain situation.

In a recent paper (\cite{erdogan2025toma}) it is shown how a Theory of Mind can build upon some of the other aspects that have already been discussed above to create a computationally efficient Theory of Mind. E.g. if a person is a medical doctor we can assume they have knowledge about medicine, they have skills with regards to diagnosing illnesses, etc. If someone is a car sales person we can assume knowledge about cars, etc. Thus even without knowing an individual we already have a lot of information about their beliefs and reasoning capacities. We can also leverage the context in which we interact with another person. When I meet a doctor in the hospital we can assume the goal of the meeting is about some disease I might have. Thus I can expect questions and information that are related to that illness. If I trust the doctor I can accept the things they are telling me and do not have to check or question every detail (unless I do not understand something).\\
The idea of a Theory of Mind is that it combines personal experience with another agent with the more general contextual knowledge. In that way we can, for instance, derive that although in general car sales persons are not trustworthy, I have interacted with the current car sales person several times before and the outcome was very good every time and the information they gave was reliable. Thus I can trust this specific person. In these ways the Theory of Mind takes both the internal mental state of other agents into account as well as the social context of the interaction to guide the interactions. Without the Theory of Mind interactions would be longer and more items would have to be made explicit or should be verified before the interactions would lead to some intended result.

By contrast, recent foundation models show only emergent and inconsistent abilities to infer mental states, lacking explicit representations of beliefs or intentions.  
Without such structure, agentic systems struggle to interpret others’ perspectives or to make their own reasoning transparent.  
Developing formal and computational representations of theory of mind therefore remains essential for achieving safe, interpretable, and socially aware AI.

\section*{Implications and Reflection}
Together, these AAMAS foundations illustrate that agency involves far more than autonomous action: it depends on explicit reasoning structures, communication protocols, social organization, and mechanisms for trust, accountability, and shared purpose.  
The comparison between AAMAS and Agentic AI shows that many of the current challenges in autonomous and interactive systems are not entirely new, but reappear in different technical forms.  
What distinguishes the AAMAS tradition is its commitment to making agency \emph{explicit}, \emph{explainable}, and \emph{socially grounded}.  
Re-engaging with these principles provides not merely historical insight but a conceptual foundation for tackling the pressing problems of transparency, coordination, and governance in today’s agentic systems.  
By combining the adaptive capabilities of large-scale learning with the representational and normative rigor of multi-agent research, future AI systems can move toward a form of agency that is both effective and accountable—capable of autonomous reasoning and collaboration while remaining aligned with human values and institutional expectations.

Beyond these conceptual correspondences, the next step is to operationalize these principles through concrete methods, tools, and governance structures. 
Agentic systems function within socio-technical contexts where human oversight, interpretability, and trust are essential; explicit representations of reasoning and commitments can support shared mental models between humans and agents, enabling meaningful supervision rather than mere control. 
Building on AAMAS traditions of formal verification, simulation-based evaluation, and logic-based reasoning, future research should develop standardized benchmarks for autonomy, cooperation, and social compliance, aligning with ongoing initiatives such as the NIST AI Risk Management Framework, EU AI Act, and e.g. the UK AI Safety Institute.
Embedding agentic systems within institutional and normative frameworks must also address questions of power and value diversity, who defines norms, whose objectives are optimized, and how plural values are represented. 
By combining conceptual rigor with these emerging governance standards, the field can move toward hybrid architectures that are not only capable and adaptive, but also transparent, accountable, and socially legitimate.

\section*{Conclusion}
The emergence of Agentic AI marks a significant step in extending artificial intelligence from passive prediction to purposeful, interactive, and context-aware behaviour. To realize this potential, the development of agentic systems requires a clearer conceptual foundation for what it means to act, decide, and coordinate within complex socio-technical environments. Some of the major problems for agents based on foundation models are that the foundation models have learned the appropriate behaviour based on data that might contain data on actions, but rarely indicates the social context in which those actions were performed. Some of that context will be implicitly taken on in the models. Thus the agentic systems might perform good in a very standard context, but they will have problems to perform well when the context changes or an exception needs to be handled.

A second important issue is that the agentic AI is a monolithic structure in which it is difficult to distinguish the different aspects that caused a decision and and action to be taken. This is problematic when agents have to cooperate and need to form expectations on each other's behaviour. 

The research tradition of Autonomous Agents and Multi-Agent Systems (AAMAS) offers the theoretical and methodological scaffolding that can strengthen these foundations. Formal models of reasoning, interaction, organization, and governance provide principled ways to represent goals, commitments, and interdependencies, concepts essential for scalable and trustworthy autonomy. Integrating such models enhances coherence, verifiability, and social grounding, turning ``agentic AI'' from ad-hoc behavioural systems into coherent, cooperative, and accountable socio-technical actors.


\bibliography{aaai2026}


\end{document}